\documentclass[10pt, twocolumn]{article}
 \usepackage{amsmath}
 \usepackage{amsfonts}
\usepackage[bookmarks=false]{hyperref}

\usepackage{graphicx}
\providecommand{\keywords}[1]
{
  \small	
  \textbf{\textit{Keywords---}} #1
}

\begin{document}

\title{Experimental Comparison of Semi-parametric, Parametric, and Machine Learning Models for Time-to-Event Analysis Through the Concordance Index}

\author{
Camila Fernandez\\
Nokia Bell Labs France\\
\href{mailto:camila.fernandez@nokia.com}{camila.fernandez@nokia.com}
\and
Chung Shue Chen\\
Nokia Bell Labs France\\
\href{chung\_shue.chen@nokia-bell-labs.com}{chung\_shue.chen@nokia-bell-labs.com}
\and
Pierre Gaillard\\
Inria\\
\href{mailto:pierre.gaillard@inria.fr}{pierre.gaillard@inria.fr}
\and
Alonso Silva\\
Safran Tech\\
\href{mailto:alonso.silva-allende@safrangroup.com}{alonso.silva-allende@safrangroup.com}
}

\date{}

\vspace{-6cm}

\maketitle




\begin{abstract}
In this paper, we make an experimental comparison of semi-parametric (Cox proportional hazards model, Aalen additive model), parametric (Weibull AFT model), and machine learning models (Random Survival Forest, Gradient Boosting Cox Proportional Hazards Loss, DeepSurv) through the concordance index on two different datasets (PBC and GBCSG2). We present two comparisons: one with the default hyperparameters of these models and one with the best hyperparameters found by randomized search.
\end{abstract}


\bigskip\bigskip


\section{Introduction}


Time-to-event analysis originated from the idea to predict the time until a certain critical event occurs.
For example, in healthcare, the goal is usually to predict the time until a patient with a certain disease dies. Another example is maintenance where the objective is to predict the time until a component fails.
There are many other examples that are of interest to time-to-event analysis such as predicting customer churn, predicting the time until a convicted criminal reoffends, etc.
One of the main challenges of time-to-event analysis is right censoring, which means that the event of interest has only occurred for a subset of the observations, making the problem different from typical regression problems in machine learning.

In this paper, we will use two datasets to perform this analysis. The first one is about patients diagnosed with breast cancer (GBCSG2)
and the second one are patients diagnosed with primary biliary cirrhosis (PBC). For the first dataset the critical event of interest will be the recurrence of cancer while for the second one it will be the death of the patient.


In each dataset and for each sample we have an observed time that could be either the survival time or the censored time. A censored time will occur when the time of death has not been observed, and then, in this case this time corresponds to the last medical record of the patient. The censored time will be a lower bound for the survival time.


\subsection{Survival and hazard functions}

The fundamental task of time-to-event analysis is to estimate the probability distribution of time until some event of interest happens.

Consider a covariates/features vector $X$,
a random variable that takes on values in the covariates/features space $\mathcal{X}$.
Consider a survival time $T$, a non-negative real-valued random variable.
Then, for a feature vector $x\in\mathcal{X}$, our aim is to estimate the conditional survival function:
\begin{equation}
S(t|x):=\mathbb{P}(T>t|X=x),
\end{equation}
where $t\ge0$ is the time and $\mathbb{P}$ is the probability function.

In order to estimate the conditional survival function $S(\cdot|x)$,
we assume that we have access to $n$ training samples, in which for the $i$-th sample we have: 
$X_i$ the feature vector, 
$\delta_i$ the survival time indicator, which indicates whether we observe the survival time or the censoring time,
and $Y_i$ which is the survival time if $\delta_i=1$ and the censoring time otherwise.


Many models have been proposed to estimate the conditional survival function $S(\cdot|x)$.
The most standard approaches are the semi-parametric and parametric models,
which assume a given structure of the hazard function:
\begin{equation}
h(t|x):= - \frac{\partial}{\partial t} \log S(t|x).
\end{equation} 




\subsection{Concordance index} 

The concordance index, introduced by Harrell et al. (1996) in~\cite{Harrell}, is the most used performance metric for time-to-event analysis.
It measures the fraction of pairs of subjects that are correctly ordered within the pairs that can be ordered.
The highest (and best) value that can be obtained is $1$, 
which means that there is complete agreement between the order of the observed and predicted times.
The lowest value that can be obtained is $0$, which denotes a perfectly wrong model, while
a value of $0.5$ means that it is a random model.

To calculate the concordance index we first take every pair in the test set such that the earlier observed time is not censored. Then we consider only pairs $(i,j)$ such that $i<j$ 
and we also eliminate the pairs for which the times are tied unless at least one of them has an event indicator value of $1$. Next, we compute for each pair $(i,j)$ a score $C_{i,j}$ which for $Y_i \neq Y_j$ is $1$ if the subject with earlier time (between $i$ and $j$) has higher predicted risk (between $i$ and $j$), is $0.5$ if the risks are tied and $0$ otherwise. For $Y_i=Y_j$ and $\delta_i = \delta_j =1$ we set $C_{i,j} = 1$ if the risks are tied and $0.5$ otherwise. If only one of $\delta_i$ or $\delta_j$ is $1$ we set $C_{i,j}=1$ if the predicted risk is higher for the subject with $\delta = 1$ and $0.5$ otherwise.

Final we compute the concordance index as follows

\begin{equation}
    \frac{1}{|\mathcal{P}|} \displaystyle \sum_{(i,j) \in \mathcal{P}} C_{i,j},
\end{equation}
where $\mathcal{P}$ represents the set of eligible pairs $(i,j)$.

\section{Datasets description}

\subsection{German Breast Cancer Study Group dataset (GBCSG2)}

The German Breast Cancer Study Group (GBCSG2) dataset, made available by Schumacher et al.~(1994) in~\cite{Schumacher},
studies the effects of hormone treatment on recurrence-free survival time.
The event of interest is the recurrence of cancer time.
The dataset has 686 samples and 8 covariates/features:
age, estrogen receptor, hormonal therapy, menopausal status (premenopausal or postmenopausal),
number of positive nodes, progesterone receptor, tumor grade, and tumor size.
At the end of the study, there were 387 patients (56.4\%) who were right censored (recurrence-free).
In our experiments, we reserve 25\% of the dataset as testing set.

\subsection{Mayo Clinic Primary Biliary Cirrhossis dataset (PBC)}

The Mayo Clinic Primary Biliary Cirrhosis dataset, made available by Therneau and Grambsch (2000) in~\cite{Therneau},
studies the effects of the drug D-penicillamine on the survival time. 
The event of interest is the death time.
The dataset has 276 samples and 17 covariates/features: age, serum albumin, alkaline phosphotase, presence of ascites,
aspartate aminotransferase, serum bilirunbin, serum cholesterol, urine copper, edema, presence of hepatomegaly or enlarged liver,
case number, platelet count, standardised blood clotting time, sex, blood vessel malformations in the skin, histologic stage of disease,
treatment and triglycerides.
At the end of the study, there were 165 patients (59.8\%) who were right censored (alive).
In our experiments, we reserve 25\% of the dataset as testing set.

\section{Models}

\subsection{Semi-parametric model: Cox proportional hazards}

Cox in~\cite{Cox} proposes a semi-parametric model, also known as Cox proportional hazards model, to estimate the conditional survival function.
This model assumes that the log-hazard of a subject is a linear function of their $m$ static covariates/features~$h_i, i\in[m]$, and a population-level baseline hazard function $h_0(t)$ that changes over time:
\begin{equation}
h(t|x)=h_0(t)\exp\left(\sum_{i=1}^mh_i(x_i-\bar{x_i})\right).
\end{equation}

The term `proportional hazards' refers to the assumption of a constant relationship between the dependent variable and the regression coefficients.  Also, this model is semi-parametric in the sense that the baseline hazard function $h_0(t)$ does not have to be specified and it can vary allowing a different parameter to be used for each unique survival time.

\subsection{Semi-parametric model: Aalen's additive model}

Aalen's additive model, proposed by Aalen (1989) in~\cite{Aalen}, estimates the hazard function
but instead of being a multiplicative model as the Cox proportional hazards model, it is an additive model.
The hazard function estimator is the following

\begin{equation}
h(t|x)= b_0(t) + \sum_{i=1}^m b_i(t)x_i.
\end{equation}

\subsection{Parametric model: Weibull Accelerated Failure Time model (Weibull AFT)}

Consider we have two survival functions for each one of two different populations, $S_A(t)$ and $S_B(t)$ and an accelerated failure rate $\lambda$ such that $S_A(t)=S_B(\frac{t}{\lambda})$ where $\lambda$ can be modeled as a function of the covariates/features and it describes stretching out or contraction of the survival time:

\begin{equation}
\lambda(x)=\exp\left(b_0 + \displaystyle \sum_{i=1}^m b_ix_i \right)
\end{equation}

Then, we suppose a Weibull form for the survival function $S(t)$ leading us to assume

\begin{equation}
    h(t|x)= \left( \frac{t}{\lambda (x)} \right)^{\rho}
\end{equation}
where $\rho$ is an unknown parameter that must be fitted. This model is called Weibull accelerated failure time shortened as Weibull AFT model.

\subsection{Machine learning model: Random Survival Forest}

The random survival forest model, proposed by Ishwaran et al. (2008) in~\cite{Ishwaran}, is an extension of the random forest model, introduced by Breiman et al. (2001) in~\cite{Breiman}, that can take into account censoring. The randomness is introduced in two ways, first we use bootstrap samples of the dataset to grow the trees and second, at each node of the tree, we randomly choose a subset of variables as candidates for the split. The quality of a split is measured by the log-rank splitting rule. Then, we average the trees results which allows us to improve the accuracy and avoid overfitting.

\subsection{Machine learning model: Random Survival Forest and Adaptative Nearest Neighbors}

We also consider a random survival forest variation from Chen (2019) in~\cite{Chen}. Each leaf will be associated to a different subset of the data set for which a Kaplan Meier survival estimator is applied, and so, each leaf is associated to a survival function estimate. Then, for a test point $x$ we choose all the leaves that $x$ belongs to and we only average the results of these leaves to obtain our final estimation.


\subsection{Machine learning model: Gradient Boosting Cox Proportional Hazards Loss}

The idea of gradient boosting was originated by Breiman and later developed by J.H. Friedman (2001) in~\cite{Friedman}. Gradient boosting is an additive model in which at each step it adds a new weak learner so that it minimizes a loss function. The model has principally three components, the loss function, the weak learner and the additive model. The loss function we aim to minimize will be the negative Cox's log partial likelihood, as proposed by Ridgeway (1999) in~\cite{Ridgeway}. At each step $i$ we have an estimator $\hat{S_i}$ and we add an estimator $\hat{h}$ which will be originated by a decision tree and such that minimizes the loss function. Then, our estimator at the stage $i+1$ will be

\begin{equation}
 \hat{S}_{i+1} (t|x) = \hat{S_i}(t|x) + \hat{h}(t|x).   
\end{equation}


\subsection{DeepSurv}

DeepSurv, proposed by Katzman et al. (2018) in~\cite{Katzman}, is a nonlinear version of Cox proportional hazards model. Cox proportional hazards is a semiparametric model that calculates the effects of observed covariates on the risk of an event occurring and it supposes that this risk is a linear combination of the covariates. However, this linear assumption may be too simplistic and not accurate enough. DeepSurv proposes to use deep neural networks to learn a nonlinear relationship between covariates/features and an individual's risk of failure.
DeepSurv is a multi-layer perceptron and it estimates for each feature $x$ the risk function $\hat{r}_{\theta}(x)$ parametrized by the weights of the network $\theta$. This function is the same function $ \displaystyle \sum_{i=1}^mh_i(x_i-\bar{x_i})$  presented in the Cox proportional hazard model, but the difference is that in this case it is not assumed to be linear and it is given by minimizing the loss function of the neural network

\begin{equation}
    l(\theta) = - \frac{1}{N} \sum_{i:\delta_i=1} \left( \hat{r}_{\theta}(x_i)- \log \sum_{j \in \mathcal{R}(Y_i)} e^{\hat{r}_{\theta}(x_j)} \right) + \lambda || \theta||^2,
\end{equation}
where $\lambda$ is a regularization parameter, $N$ is the number of uncensored subjects and  $\mathcal{R}(t)$ is the set of subjects at risk at time $t$.

\section{Results and Conclusions}

We compared all the models described for the two datasets through the concordance index. To do this analysis we used Scikit-learn~\cite{Pedregosa}, Lifelines~\cite{Davidson}, Scikit-survival~\cite{Polsterl}, and Matplotlib~\cite{Hunter}.
For each dataset, the experiment we performed is the following: we choose $25$ different seeds for splitting the dataset, this generates $25$ different partitions between training and validation sets. Then we run the model $25$ times (one for each partition) and we make a boxplot with the distribution of the concordance indexes obtained. In the figures, we can observe the median of the obtained concordance indexes represented by the red lines and the average represented by the red triangles.

Figure 1 shows the comparison of the concordance indexes for PBC dataset where we can appreciate that random survival forest model fitted with a random search of the hyperparameters outperforms the other models. Figure 2 shows the comparison of the concordance indexes for GBCSG2 dataset and we can see that random survival forest with adaptive nearest neighbors outperforms the other models.

Furthermore, we can observe that traditional methods performed reasonably well for the small dataset PBC (see Cox proportional hazards with randomized search), but they underperformed against machine learning methods for the larger dataset (GBCSG2). We can also observe that the deep learning method (Deepsurv) did not perform better than random survival forest model and therefore the progress made by deep learning in other areas (computer vision, NLP, etc.) has not yet been replicated for time-to-event analysis.

Classical methods for predicting survival time are easier to interpret and to analyze the way in which each covariate/feature has an influence in the model. For the case of PBC dataset, random survival forest with random search outperforms Cox proportional hazards with random search by less than $1\%$ while in GBCSG2 the method RSF with adaptive nearest neighbor increase the performance by $2.5\%$ with respect to randomized search Cox proportional hazards model. Therefore, if this increment in performance is significant enough to compensate for the loss of easier interpretation of the model will highly depend on the application.

\section*{Acknowledgments}
The work presented in this paper has been partially carried out at LINCS (\url{http://www.lincs.fr}).





\onecolumn

\begin{figure}
\begin{center}
\includegraphics[height=7.5cm,width=12cm]{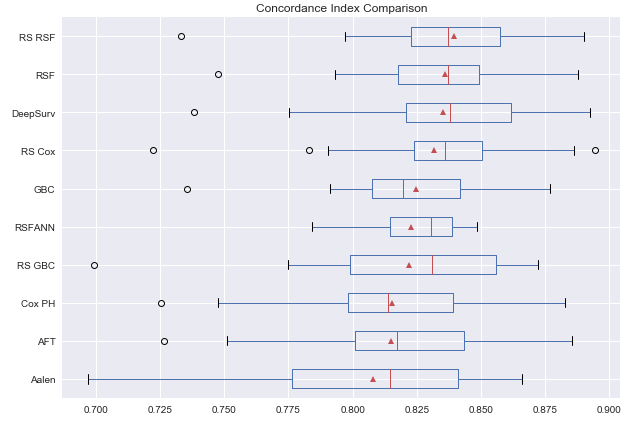}
\caption{Concordance index comparison for PBC dataset.}
\label{graph1}
\end{center}


\begin{center}
\includegraphics[height=7.5cm,width=12cm]{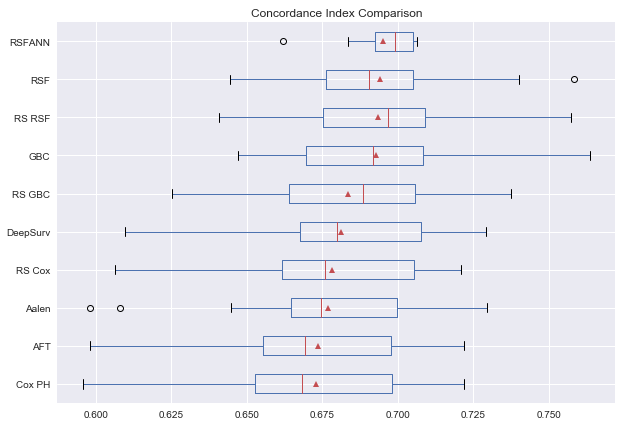}
\caption{Concordance index comparison for GBCSG2 dataset.}
\label{graph1}
\end{center}
\end{figure}

\twocolumn

\end{document}